\documentclass{article}

\usepackage{arxiv}
\usepackage[utf8]{inputenc} 
\usepackage[T1]{fontenc}    
\usepackage{url}            
\usepackage{booktabs}       
\usepackage{amsfonts}       
\usepackage{nicefrac}       
\usepackage{microtype}      
\usepackage{hyperref}
\usepackage{graphicx}
\setcitestyle{authoryear,round,citesep={;},aysep={,},yysep={;}}
\usepackage{doi}
\usepackage[table,svgnames]{xcolor}
\usepackage{capt-of}
\newcommand{\sys}{TweezeEdit\xspace}
\definecolor{skyblue}{RGB}{101,162,217}
\definecolor{main}{HTML}{777777} 
\definecolor{sub}{HTML}{bbbbbb}
\usepackage{pifont}
\usepackage{amsmath}
\usepackage{booktabs}
\usepackage{multirow}
\usepackage{multicol}
\def\secref#1{Section~\ref{#1}}
\def\figref#1{Figure~\ref{#1}}
\def\eqref#1{Eq~\ref{#1}}
\newcommand{\tabref}[1]{Table \ref{#1}}
\def\algref#1{Algorithm~\ref{#1}}
\usepackage[most]{tcolorbox}
\newtcolorbox{boxE}{
    enhanced, 
    boxrule = 0pt, 
    colback=white,
    borderline = {0.75pt}{0pt}{main}, 
    borderline = {0.75pt}{2pt}{sub} 
}
\usepackage[ruled]{algorithm2e}
\usepackage{wrapfig}
\usepackage{cleveref}       

\title{\sys: Consistent and Efficient Image Editing with Path Regularization}
\usepackage{authblk}

\author{%
	Jianda Mao
}
\author{%
	Kaibo Wang
}
\author{%
	Yang Xiang
}
\author{%
	Kani Chen\thanks{\texttt{Correspondence to: Kani Chen <makchen@ust.hk>.}}
}
\affil{Department of Mathematics, The Hong Kong University of Science and Technology, Hong Kong SAR}

\begin{document}
\maketitle

\begin{figure*}[h]
  \centering
  \includegraphics[width=0.95\textwidth]{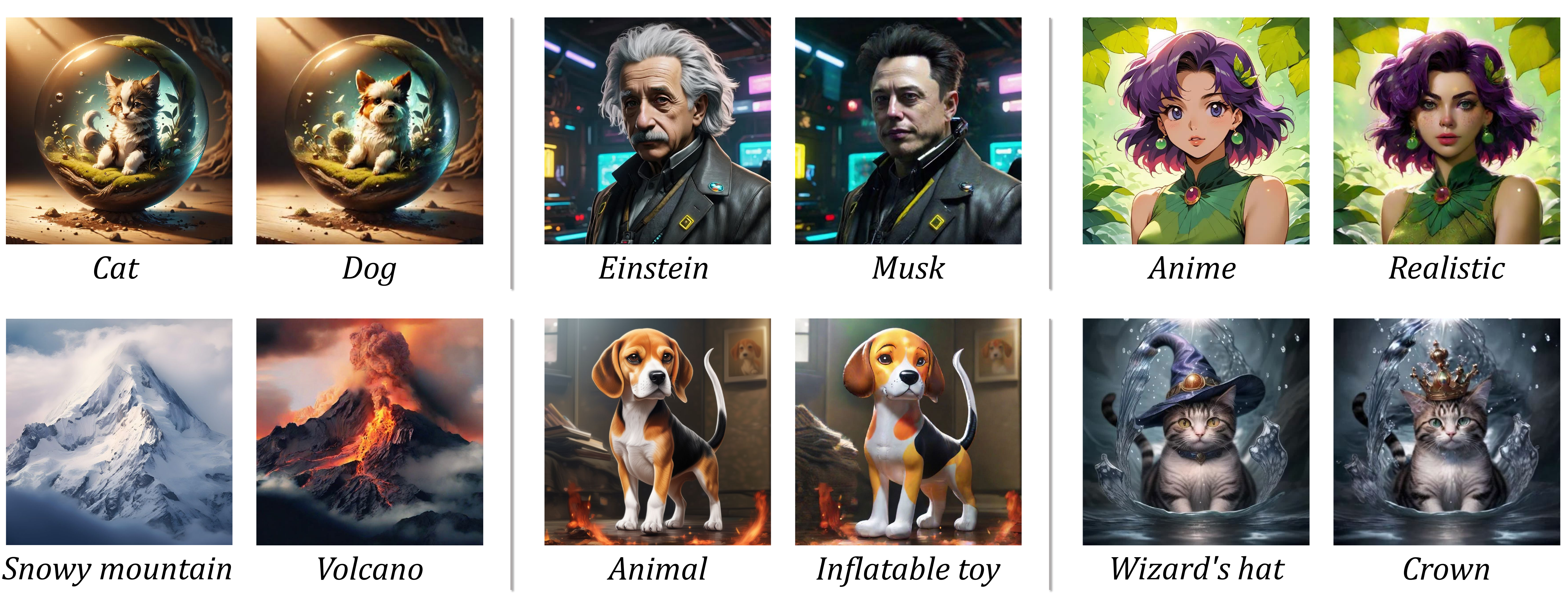}
  \caption{Examples of images edited with \textit{TweezeEdit}, a tuning- and inversion-free framework for text-driven editing using pretrained consistency models. It effectively preserves source image semantics while aligning with target prompts.}
  \label{fig:teaser}
\end{figure*}

\begin{abstract}
Large-scale pre-trained diffusion models empower users to edit images through text guidance. However, existing methods often over-align with target prompts while inadequately preserving source image semantics. Such approaches generate target images explicitly or implicitly from the inversion noise of the source images, termed the inversion anchors. We identify this strategy as suboptimal for semantic preservation and inefficient due to elongated editing paths. We propose \textit{TweezeEdit}, a tuning- and inversion-free framework for consistent and efficient image editing. Our method addresses these limitations by regularizing the entire denoising path rather than relying solely on the inversion anchors, ensuring source semantic retention and shortening editing paths. Guided by gradient-driven regularization, we efficiently inject target prompt semantics along a direct path using a consistency model. Extensive experiments demonstrate TweezeEdit’s superior performance in semantic preservation and target alignment, outperforming existing methods. Remarkably, it requires only 12 steps (1.6 seconds per edit), underscoring its potential for real-time applications.
\end{abstract}

\section{Introduction}
Recent advances in large-scale diffusion models~\cite{rombach2022high,luo2023latent,flux2024} have enabled text-driven image editing~\cite{xu2024inversion, rout2024semantic, kulikov2024flowedit,hertz2022prompt}, where a source image and target prompt are input to produce an edited image. However, existing methods often fail to preserve the semantic content of the source image, over-aligning with the target prompt and necessitating extensive corrections with additional control~\cite{shuai2024survey}. Current approaches typically employ a deterministic reverse process to derive \textit{inversion anchors} (the inverted noise of source images). Although inversion anchors theoretically encapsulate source image information, they often fail to generate expected similar denoising paths between source and target images in practice, which results in over-alignment during editing~\cite{mokady2023null, miyake2023negative}.

Two primary issues contribute to this challenge: (1) numerical errors in inversion cause information loss, preventing the inversion anchors from fully reconstructing the source image, and (2) the lack of constraints on the diffusion model’s output results in uncontrolled divergence in the denoising path of source and target images from the inversion anchors, triggering unexpected changes. Although recent efforts~\cite{kulikov2024flowedit} mitigate inversion errors by interpolating direct paths between source and target images using sampled inversion anchors, they still suffer from inaccuracies in estimation and inherent over-alignment from the inversion paradigm. Moreover, techniques~\cite{cao2023masactrl, hertz2022prompt} that impose path constraints often require intrusive modifications, which rely on model-specific designs and increase resource demands. Tuning-based methods~\cite{mokady2023null, zhang2023forgedit} align output with the source image by tuning text embeddings or parameters, yet they require heavy computation and may overfit, harming general generative ability.

To address these limitations, we propose \textit{TweezeEdit}, a tuning-free, inversion-free framework for efficient and semantically consistent image editing. Unlike methods that depend solely on inversion anchors, our approach regularizes the entire denoising path difference between the source and target images. This strategy constrains the output of the diffusion model, akin to tightening the arms of tweezers, restricting edits exclusively to prompt-relevant regions. Our method not only enhances source retention but also shortens the direct path. Additionally, we employ gradient-based regularization to guide updates along this path, eliminating the need for architectural modifications. Using consistency models as the backbone, TweezeEdit reduces sampling steps and cumulative errors for efficient editing, while naturally extending to noise and velocity prediction models through their connection to consistency models.

Through extensive experiments, we demonstrate that TweezeEdit outperforms state-of-the-art (SOTA) methods in prompt alignment and semantic preservation. As an architecture-agnostic solution, TweezeEdit seamlessly integrates with attention control for refinement. Quantitative and qualitative evaluations highlight its superiority in consistency-critical tasks and perceptual quality. Leveraging consistency models, TweezeEdit achieves edits in approximately 12 sampling steps, reducing latency while maintaining quality.

Our contributions are threefold:  
\begin{enumerate}
    \item We extend the inversion anchor paradigm by regularizing the entire denoising path, enhancing source semantic retention and shortening editing paths.
    \item We propose \textit{TweezeEdit}, a gradient-guided editing algorithm that avoids inversion and architectural changes, and is accelerated by consistency models.
    \item We empirically validate TweezeEdit’s effectiveness and efficiency in editing tasks.
\end{enumerate}

\section{Related Work}
In diffusion-based image editing, starting from a source image and its description (source prompt), the central challenge lies in modifying the image according to a target prompt while preserving consistency with the source image.

\textbf{Tuning-based methods}~\cite{ruiz2023dreambooth, dong2023prompt, zhang2023forgedit, mokady2023null} enforce the models to reconstruct the source image given the source prompt by optimizing text embeddings or model parameters. While effective for consistency, these approaches are computationally intensive and may compromise the models' generative capability due to overfitting.

\textbf{Tuning-free methods} leverage pre-trained diffusion models for image editing without fine-tuning. These approaches rely on inversion anchors (i.e., inverted noise from source images) to preserve structural and semantic consistency, but they face challenges in reconstruction fidelity due to inversion errors. For example, DDIM~\cite{song2020denoising} suffers from error accumulation~\cite{mokady2023null, miyake2023negative} during its reverse denoising process when estimating the initial noise. RF-inversion~\cite{rout2024semantic} improves consistency by integrating conditional vector fields based on source images, but it remains limited by inversion inaccuracies. Moreover, direct modifications to these vector fields may degrade generation quality or introduce semantic distortions. FlowEdit~\cite{kulikov2024flowedit} employs sampling-based inverted noise approximation to avoid explicit inversion, yet it still struggles with anchor precision. Virtual Inversion~\cite{xu2024inversion} generates stochastic paths for source and target images and calibrates the target paths using reconstruction errors from the source paths. However, this method requires additional constraints due to the stochasticity of the paths.

\textbf{Attention-based methods}, a subset of tuning-free methods, use diffusion models' attention mechanism to guide image editing. In U-Net-based models~\cite{rombach2022high}, some approaches~\cite{hertz2022prompt,cao2023masactrl} refine attention during generation by leverage attention maps from the source image reconstruction. StableFlow~\cite{avrahami2024stable} extended these to transformer-based models~\cite{peebles2023scalable}, improving source-target consistency. However, these methods require architectural changes and higher computation, reducing scalability and efficiency.

\section{Preliminaries}
\subsection{Diffusion Models}
Diffusion models~\cite{ho2020denoising} are generative models that learn to reverse a gradual noising process to produce images. In practice, diffusion models typically operate in a latent space~\cite{rombach2022high}, obtained by encoding images through an encoder-decoder architecture. For notational simplicity, we formulate the diffusion process as acting directly on the image $z_0$.

During forward diffusion ($t = 0 \to +\infty $, discretized in practice as $t\in\{0,1,\cdots,T\}$), the clean image $z_0$ is progressively corrupted into Gaussian noise. The noisy image  $z_t$ at timestep  $t$ follows:
\begin{equation}
    z_t=\sqrt{\bar\alpha_t}z_0+\sqrt{1-\bar\alpha_t}\epsilon, \quad\epsilon\sim\mathcal{N}(0,I),
\end{equation}
where $\alpha_{1:T} $ and  $\bar\alpha_t = \prod_{s=1}^t \alpha_s$ are schedule parameters.
When $T \to \infty $,  $\bar\alpha_T \rightarrow 0 $ ensures  $z_T\rightarrow \mathcal{N}(0,I) $.

Following DDIM~\cite{song2020denoising}, the reconstruction process reverses the forward diffusion via iterative denoising according to:
\begin{equation}
\label{eq: denoise}
        z_{t-1}\!=\!\sqrt{\bar\alpha_{t\!-\!1}}(\frac{z_t\!-\!\sqrt{1\!-\!\bar\alpha_t}\epsilon_\theta(z_t,t)}{\sqrt{\bar\alpha_t}})
        \!+\! \sqrt{1\!-\!\bar\alpha_{t-1}} \epsilon_\theta(z_t,t),
\end{equation}
starting from  $z_T \sim \mathcal{N}(0,I)$, where $\epsilon_\theta(z_t, t)$ is a trained noise-prediction network output. This discrete update corresponds to a numerical solver for a deterministic ODE~\cite{song2020score}.
\begin{wrapfigure}{r}{0.48\textwidth}
  \centering
  \includegraphics[width=0.48\textwidth]{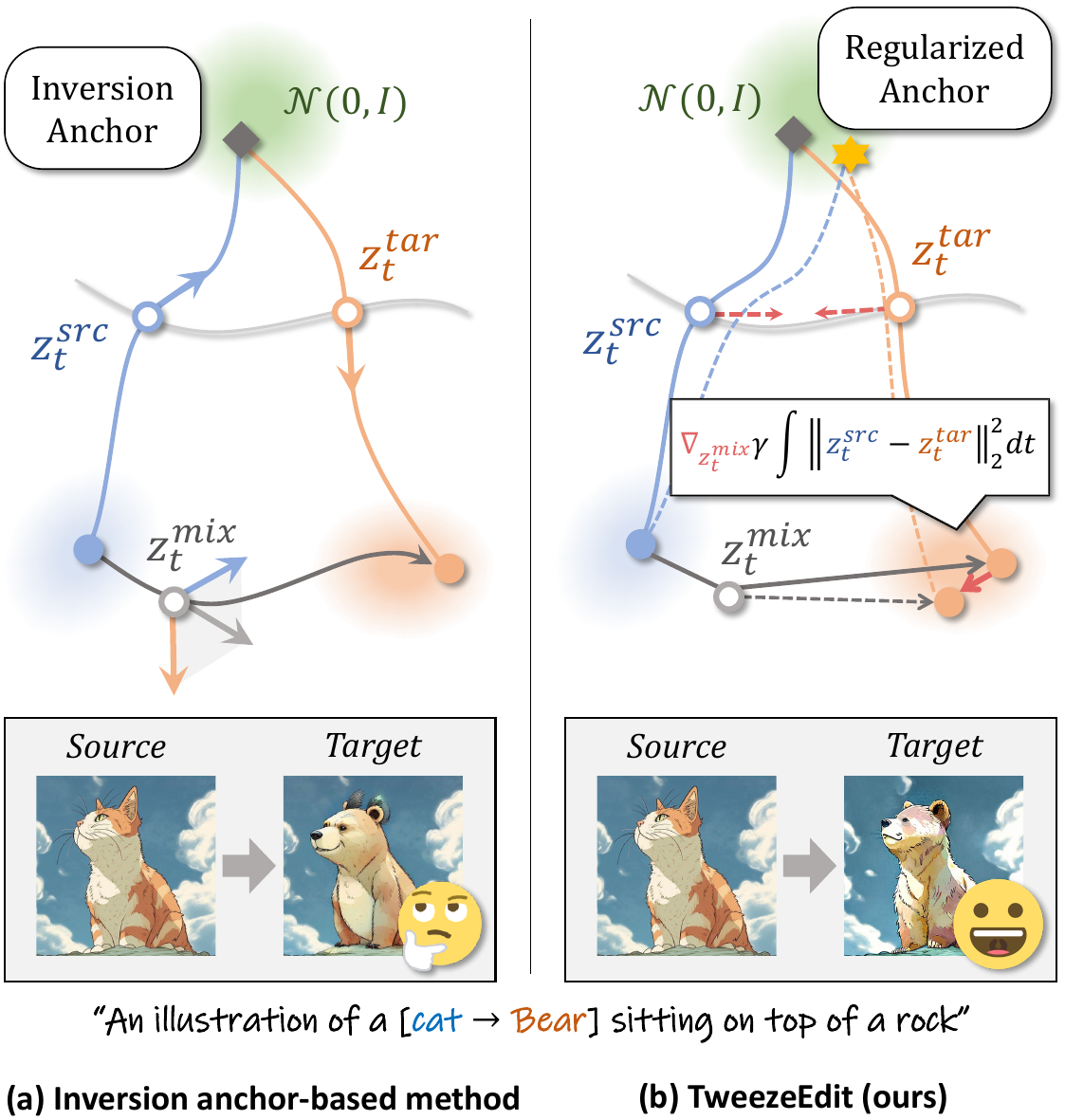}
  \captionof{figure}{Comparison of inversion anchor-based method (a) and \sys (b). The inversion anchor-based method follows either the inversion-denoising path (\textcolor[RGB]{143,170,220}{\ding{220}}\textcolor[RGB]{118,113,113}{\ding{117}}\textcolor[RGB]{244,177,131}{\ding{220}}) or equivalent direct path (\textcolor[RGB]{118,113,113}{\ding{220}}), frequently over-aligning with the target prompt due to inadequate retention of source semantics. \sys navigates the direct path via a consistency model, implicitly calibrating the anchor through the gradient (\textcolor[RGB]{227,99,99}{\ding{220}}) of denoising path regularization. Our method tightens subsequent denoising paths (shown as dashed lines), producing target images that better preserve source content.}
  \label{fig: method}
      \vspace{-30pt}
\end{wrapfigure}

\subsection{Consistent Models}
Consistency models~\cite{song2023consistency,luo2023latent}, a class of diffusion models, enhance sampling efficiency by enforcing self-consistency across timesteps. These models learn a mapping $f(z_t, t)$ that directly predicts the clean image $z_0$ from a noisy input $z_t$. In practice, they employ multistep consistency sampling to indirectly refine $z_0$:  
\begin{equation}  
\label{eq:cm-sampling}  
    \hat{z}_{t-k} = \sqrt{\bar\alpha_{t}}f_\theta(\hat{z}_{t}, t) + \sigma_{t}\epsilon, 
\end{equation}  
where $\epsilon \sim \mathcal{N}(0,I)$, and $\sigma_{t}$ denotes the noise scale at timestep $t$. By allowing larger step sizes $k$, consistency models generate images with fewer sampling steps.

\subsection{Tuning-Free Image Editing}
Tuning-free image editing methods based on diffusion models involve a two-step process: inversion followed by denoising. During inversion, the source image $z_0^{src}$ is inverted into noise (inversion anchors) either explicitly via DDIM inversion or implicitly via sampling ~\cite{kulikov2024flowedit} (i.e., $z_0^{src}\rightarrow z_T$). In the subsequent denoising phase, guided by a target prompt, $z_T$ is denoised to generate the target \ image $z_0^{tar}$, aiming to preserve consistency with the source image while aligning the output with the target prompt.

Some methods ~\cite{kulikov2024flowedit} construct an update target $z_t^{mix}=z_0^{src}+z_t^{tar}-z_t^{src}$, integrating semantics from $z_t^{src}$ into the update process to enhance consistency, rather than relying solely on $z_T$. This mathematically corresponds to a direct path between $z_0^{src}$ and $z_0^{tar}$. 
Other approaches~\cite{tumanyan2023plug,cao2023masactrl,hertz2022prompt} enforce consistency through attention control of diffusion models.

\section{TweezeEdit}
Current methods depend on inversion anchors to maintain consistency with source images. However, inaccurate estimation of these anchors often results in poor consistency preservation. To address the dual challenges of preserving source image semantics while achieving alignment with target prompts, we propose \textit{\sys}, a consistent and efficient image editing framework. Our approach builds on a key intuition illustrated in \figref{fig: method} (with corresponding elements labeled in parentheses):

\begin{boxE}
    \textit{\underline{Key Intuition}}:\\
    \ding{118} \textbf{Desired edited image}: The desired edited image (\textcolor[RGB]{244,177,131}{\ding{108}}) in the target prompt’s distribution stays close to the source image (\textcolor[RGB]{143,170,220}{\ding{108}}), ensuring both semantic consistency and prompt alignment.
    
    \ding{118} \textbf{Desired anchor}: Source and desired edited images derive from an identical noise (regularized anchor, \textcolor[RGB]{255,192,0}{\ding{72}}) with a tightened denoising path (\textcolor[RGB]{143,170,220}{\ding{220}}\textcolor[RGB]{255,192,0}{\ding{72}}\textcolor[RGB]{244,177,131}{\ding{220}}) instead of source image's DDIM-inverted noise (inversion anchor, \textcolor[RGB]{118,113,113}{\ding{117}}).
\end{boxE}

Based on this principle, \sys integrates two core components. First, we leverage a consistency model to progressively integrate target semantics along the direct path (\secref{sec: conmodel}). By reducing sampling steps and incorporating a calibration trick, the model mitigates cumulative errors and enhances alignment. Second, we introduce denoising path regularization (\secref{sec: regular}), which enhances semantic preservation by constraining the entire denoising path, rather than solely depending on the inversion anchors. This approach additionally shortens the direct path - akin to tweezers tightening mechanical arms, allowing the regularization to sharpen editing focus while maintaining source fidelity.

\subsection{Direct Path with Consistency Model}
\label{sec: conmodel}
\textbf{Direct path interpolation.} To inject the target prompt into the source image, we construct an interpolation path between them using consistency models. We first define the ideal direct path between the source and target images as:
\begin{equation}
z^{mix}_t = z_0^{src} + \sqrt{\bar{\alpha}_t}(z_0^{tar} - z_0^{src}),
\end{equation}
where $\bar{\alpha}_t$ is the diffusion noise scheduler (with $\bar{\alpha}_0 = 1$ and $\bar{\alpha}_T = 0$), and $z^{mix}_t (t =1,\cdots,T)$ represents the interpolation that satisfies $z_T^{mix}=z_0^{src}$ and $z_0^{mix}=z_0^{tar}$.

Since $z_0^{tar}$ is unavailable, we approximate $z_t^{mix}$ using:

\begin{equation}
    \label{eq: iter}
    z^{mix}_t = z_0^{src} + \sqrt{\bar{\alpha}_t}(f(z_{t+1}^{tar},t+1) - z_0^{src}),
\end{equation}
Notably, $z_t^{src}$ and $z_t^{tar}$ are derived from the diffusion process with shared noise $\epsilon$, i.e., $z_t^{src/tar} = \sqrt{\bar{\alpha}_t}z_0^{src/tar} + \sqrt{1-\bar{\alpha}_t}\epsilon$. Theoretically, shared noise minimizes the upper bound of the distance between $z_t^{src}$ and $z_t^{tar}$, while empirically it prevents noisy artifacts (please see Appendix ~\ref{app: shared} for details). For notational brevity, $src/tar$ denotes that the equation applies to both cases. This shared noise formulation yields the key relationship $\sqrt{\bar{\alpha}_t}(z_0^{tar} - z_0^{src}) = z_t^{tar} - z_t^{src}$, allowing  $z_t^{mix}$ can be equivalently expressed as:
\begin{equation}
\label{eq: tmix}
z^{mix}_t = z_0^{src} + z_t^{tar} - z_t^{src}.
\end{equation}
The evolution of $z_t^{src}$ and $z_t^{tar}$ enables the consistency model to iteratively refine $f(z_t^{tar},t)$, yielding progressively better approximations of $z_0^{tar}$ for direct path simulation.

\textbf{Consistency model-based editing.} A straightforward approach is to sample $z_t^{src}$ from a given $z_0^{src}$ and represent $z_t^{tar}$ as $z^{mix}_t - z_0^{src} + z_t^{src}$ using \eqref{eq: tmix}. The denoised $f(z_{t}^{src},t)$ and $f(z_{t}^{tar},t)$ are then obtained from the consistency model, and $z^{mix}_{t-1}$ is updated according to \eqref{eq: iter}, iterating until $z^{mix}_{0} = z^{tar}_0$. 

However, this approach faces two challenges: (1) Sampled $z_t^{src}$ may not match the desired anchor. We address this through on-the-fly noise rectification via denoising path regularization, as detailed in \secref{sec: regular}. (2) Update errors in $f(z_{t}^{tar},t)$ can degrade alignment with the target prompt. \sys addresses this issue by employing a consistency model as the denoising algorithm, effectively reducing cumulative errors and enabling the calibration trick for improved target alignment.

When $z_0$ is known, the prediction error of $f(z_t^{src},t)$ can be accurately derived. Building on this,
we enhance the precision of $f(z^{tar}_t, t)$, by applying the \textbf{calibration trick} from~\cite{ju2023direct}, adjusting the prediction as:
\begin{equation}
    \label{eq: approx}
    \hat{f}(z^{tar}_t, t) = f(z^{tar}_t, t) + z_0^{src} -f(z^{src}_t, t),
\end{equation}
which calibrates $f(z^{tar}_t, t)$ by leveraging the denoising path similarity between the source and target. The prediction $\hat{z}_{t-1}^{mix}$ for $z_{t-1}^{mix}$ then becomes:
\begin{equation}
\begin{aligned}
    \hat{z}_{t-1}^{mix} &= z^{src}_0+\sqrt{\bar{\alpha}_{t-1}}( \hat{f}(z^{tar}_t, t)-z^{src}_0)\\
    &= z^{src}_0+\sqrt{\bar{\alpha}_{t-1}}(f(z^{tar}_t, t)  -f(z^{src}_t, t)).
\label{eq: ref}
\end{aligned}
\end{equation}

Using \eqref{eq: ref}, we can progressively edit $z_t^{mix}$ from $z_0^{src}$ to $z_0^{tar}$ via consistency models. Compared to other diffusion models, the consistency model offers two advantages: (1) Fewer sampling steps and self-consistency constraints help mitigate potential cumulative error, while not suffering from the accumulated errors introduced during the DDIM inversion~\cite{mokady2023null, miyake2023negative}. The stability of the consistency model allows for larger step sizes ($t \to t-k$) and fewer iterations, thereby improving editing efficiency. (2) Unlike diffusion models that predict noise or velocity, the consistency model is inherently robust to noise, allowing it to perform denoising across various noise levels~\cite{song2023consistency}. This property aligns naturally with our on-the-fly noise regularization framework. 

\begin{algorithm}[tt]
\caption{TweezeEdit}
\label{alg}
    \SetKwInOut{Input}{Input}
    \SetKwInOut{Output}{Output}
    \Input{Source image $z_0^{src}$, source prompt $P^{src}$, target prompt $P^{tar}$, regularization scheduler $\hat{\gamma}_t$ and consistency model $f$}
    \Output{Edited image $z_0^{tar}$}
    $z_T^{mix} = z_0^{src}$ {\textcolor{gray}{// initialization}} \; 
    \For{$t \leftarrow T$ \KwTo $1$}{
        \textit{\textcolor{skyblue}{ Obtain samples in denoising path}}\\
        Sample $\epsilon\sim\mathcal{N}(0,I)$\;
        $z_t^{src}=\sqrt{\bar{\alpha}_t}z_0^{src}+\sqrt{1-\bar{\alpha}_t}\epsilon$\;
        $z_t^{tar}=z_t^{mix}-z_0^{src}+z_t^{src}$\;
        \textit{\textcolor{skyblue}{ Consistency model's prediction}}\\
        $\hat{z}_0^{src}=f(z_t^{src},t,P^{src})$\;
        $\hat{z}_0^{tar}=f(z_t^{tar},t,P^{tar})$\;
        \textit{\textcolor{skyblue}{Editing direction in direct path (}}\\
        $v_t=z_0^{src}+\sqrt{\bar{\alpha}_{t-1}}(\hat{z}_0^{tar}-\hat{z}_0^{src})$  \textcolor{gray}{// \eqref{eq: ref}}\;
        \textit{\textcolor{skyblue}{ Gradient of path regularization}}\\
        $\nabla_{z_t^{mix}}R_t =   \hat{\gamma}_t \left[z_t^{src} - z_t^{tar} - \frac{\dot{\bar{\alpha}}_t}{4\sqrt{\bar{\alpha}_t}} (\hat{z}_0^{src} - \hat{z}_0^{tar})\right]$ \textcolor{gray}{// \eqref{eq: grad}}\;
        \textit{\textcolor{skyblue}{ Update step}}\\
        $z_t^{mix}=v_t-\nabla_{z_t^{mix}}R_t$ \textcolor{gray}{// \eqref{eq: update}}\;}
    \Return Edited image $z_0^{tar}=z_0^{mix}$
\end{algorithm}

\subsection{Denoising Path Regularization}
\label{sec: regular}
\textbf{The Role of Inversion Anchors.} Inversion anchors serve to maintain consistency between the edited image and the source image. However, their effectiveness in preserving source semantics relies on two assumptions that often fail in practice: 
\begin{enumerate}
    \item \textit{Inversion anchors completely preserve information from the source image.} While theoretically valid, in practice, the estimation of the inversion anchors suffers from discretization and approximation errors, leading to information loss.
    \item \textit{Diffusion model updates ($t \to t-1$) modify only regions corresponding to prompt differences for semantic consistency between $z_t^{src}$ and $z_t^{tar}$.} In reality, updates uncontrollably modify existing elements, e.g., introducing undesired demeanor or body shape changes in \figref{fig: method} (a).
\end{enumerate}

In summary, inversion anchors expect semantic consistency in paired $(z_T^{src}, z_T^{tar}) \cdots (z_t^{src}, z_t^{tar}) $ $\cdots (z_0^{src}, z_0^{tar})$, which is essentially the similarity between the denoising path $z_T^{src} \cdots z_0^{src}$ and $z_T^{tar} \cdots z_0^{tar}$. Thus, rather than relying solely on inaccurate inversion anchors, regularizing the entire denoising path's similarity is more effective.

\textbf{Denoising path regularization.} We incorporate the distance of continuous denoising paths as a regularization term during the update of $z_t^{mix}$ in \eqref{eq: ref}, defined as
\begin{equation}
    R_t = \gamma_t \int_{t-1}^t \lVert z_{\tau}^{src} - z_{\tau}^{tar} \rVert_2^2 \ \mathrm{d}\tau,
\end{equation}
where $\gamma_t$ denotes the predefined regularization strength at step $t$. 

To approximate the continuous-time integral, we apply the integral mean value theorem with a Taylor expansion and derive the gradient regularization term for $z_t^{mix}$ (please see Appendix~\ref{app: gradient} for details):
\begin{equation}
\label{eq: grad}
 \nabla_{z_t^{mix}} R_t \approx  \hat{\gamma}_t \left[ z_t^{src} - z_t^{tar} - \frac{\dot{\bar{\alpha}}_t}{4\sqrt{\bar{\alpha}_t}} (f(z_t^{src}, t) - f(z_t^{tar}, t))\right],
\end{equation}
where we set $\Delta_t \approx \frac{1}{2}$ and define $\hat{\gamma}_t := 2\gamma_t \big(-1 + \frac{\dot{\bar{\alpha}}_t}{4{\bar{\alpha}_t}}\big)$ for simplicity.
Further details regarding the choice of $\hat{\gamma_t}$ can be found in Appendix ~\ref{app: choice}.

Finally, the update of $z_t^{mix}$ shown in \eqref{eq: update} can be divided into two parts: the direction of editing toward the target prompt and the direction of preserving the semantics of the source image. Our algorithm is summarized in \algref{alg}. We set the update interval to 1 only for clarity of exposition. Our method naturally scales to larger intervals. In practice, we select only 12-15 timesteps in $\{1,\cdots, T\}$, which dramatically reduces computational steps while maintaining performance.
\begin{equation}
\label{eq: update}
 \hat{z}_{t-1}^{mix} = \underbrace{z^{src}_0+\sqrt{\bar{\alpha}_{t-1}}(f(z^{tar}_t, t)  -f(z^{src}_t, t))}_{\text{target editing}}\underbrace{-\nabla_{z^{mix}_t} R_t}_{\text{source preserving}}.
\end{equation} 

By incorporating the regularization term  $R_t $ at each update step, we enforce consistency across the full denoising path as $\sum_{t=1}^T R_t = \int_{t=0}^T\gamma_{\tau}\lVert z_\tau^{src}- z_\tau^{tar}\rVert_2^2 \mathrm{d} \tau$. This requires $z_t^{tar}$ to retain source image semantics and introduces only target prompt-related changes to avoid gradient penalties. Unlike inversion anchors, our approach extends regularization across the entire denoising path and dynamically calibrates the regularized anchors, enhancing semantic consistency.  

Besides, our approach does not explicitly compute anchors. Instead, gradient-driven updates guide $z_t^{mix}$ along the direct path, which circumvents two limitations: (1) \textbf{Reduced editing path length.} Traditional inversion anchors require computationally intensive and imprecise DDIM inversion, limiting efficiency. Even when anchors are estimated via sampling~\cite{kulikov2024flowedit}, inconsistent denoising paths incur longer direct paths. By regularizing the denoising paths, we shorten the direct path. (2) \textbf{Architecture-agnostic updates.} Prior methods often compensate for semantic loss through architecture-specific interventions (e.g., attention injection). Our gradient-driven regularization operates without model intrusion, ensuring compatibility across diverse architectures and reducing computational overhead.

Although our method is based on consistency models, which offer advantages such as self-consistency and fewer denoising steps, it can still be applied to other noise or velocity prediction models due to the inherent relationship between clean-image prediction and their outputs (details are provided in ~\ref{app: para}). Meanwhile, continuous consistency models~\cite{lu2024simplifying} allows distilling these models into consistency models.
\section{Experiments}

\begin{table*}[t]
\centering
\setlength{\tabcolsep}{1mm}
\fontsize{8.5}{9.5}\selectfont
  \begin{tabular}{l c c c c c c c c c c}
    \toprule
    \multirow{2}{*}{Method} & {Structure} & \multicolumn{4}{c}{Unedited Region Preservation} & \multicolumn{2}{c}{Editing Alignment} & \multicolumn{2}{c}{Efficiency}\\
    & Distance$_{10^3}\downarrow$ & PSNR $\uparrow$ & LPIPS$_{10^3}\downarrow$ & MSE$_{ 10^4}\downarrow$ & SSIM$_{10^2}\uparrow$ & Whole $\uparrow$ & Edited $\uparrow$ & Inv-Free & Steps$\downarrow$ \\
    \midrule
    \rowcolor{gray!15}
    DDIM (SD1.5) & 79.54 & 17.36 & 220.13 & 243.13 & 70.52 & 27.08 & 23.90 & {\ding{56}} & 50 \\
    \rowcolor{gray!15}
    DDIM (SD1.5) + P2P & 69.99 & 17.87 & 208.90 & 219.56 & 71.63 & 25.28 & 22.57  & {\ding{56}} & 50\\

    \rowcolor{blue!10}
    FlowEdit (Flux) & 22.77	& 23.08	& 100.96 & 74.31 & 86.29 & 25.19 & 22.29 & \ding{52}& 28 \\
    \rowcolor{blue!10}
    RF-inversion (Flux) & 55.08 & 19.27 & 227.59 & 164.21 & 66.78 & 25.22 & 22.49 & {\ding{56}} &28	\\
    \rowcolor{blue!10}
    Stable-flow (Flux) & 16.44 & 24.24 & 76.10 & 64.41 & \textbf{89.43} & 23.98 & 20.96 & \ding{56} & 50 \\

    \rowcolor{green!10}
    VI (LCM: SD1.5)& 113.83 & 13.93 & 292.99 & 518.98 & 59.37 & \textbf{27.68} & \textbf{24.37} & {\ding{52}} & 15\\ 
    \rowcolor{green!10}
    VI (LCM: SD1.5) + P2P & 27.86 & 21.82 & 86.62 & 124.45 & 80.89 & 24.76 & 21.71 & {\ding{52}} & 15 \\

    \midrule
    \rowcolor{gray!15}
    \sys (SD1.5) & 23.96 & 22.30 & 82.62 & 83.61 & 82.11 & 25.87 & 22.45 & \ding{52}&25\\
    \rowcolor{blue!10}
    \sys (Flux) &	20.92&23.49&82.72&72.87&87.70&25.23&22.30&\ding{52}&28\\	
    \rowcolor{green!10}
    \sys (LCM: SD1.5) & 17.36 & 24.62 & 81.90 & 54.42 &80.40   & 25.54 & 22.30 & {\ding{52}} & \textbf{12} \\
    \rowcolor{green!10}
    \sys (LCM: SD1.5)+P2P & \textbf{13.63} & \textbf{25.59} & \textbf{67.36} & \textbf{43.71} & 82.65 & 24.75 &	21.61 & {\ding{52}} & \textbf{12}  \\
    \rowcolor{green!10}
    \sys (LCM: SDXL1.0) & 22.42 & 24.13 & 98.45 & 57.33 & 83.43 & 26.01 & 22.79 & {\ding{52}} & 15\\
    \bottomrule
\end{tabular}
  \caption{Quantitative results on PIE-Bench. Inv-Free indicates whether explicit inversion is avoided. Whole and Edited refer to CLIPScore for the full image and edited region. $\uparrow$: higher is better, $\downarrow$: lower is better. Bold: best results. Our approach \sys achieves competitive edits with fewer steps and high consistency in unedited areas.}
  \label{tab: quanti}
\end{table*}

\begin{figure*}[tt]
    \centering
  \includegraphics[width=\textwidth]{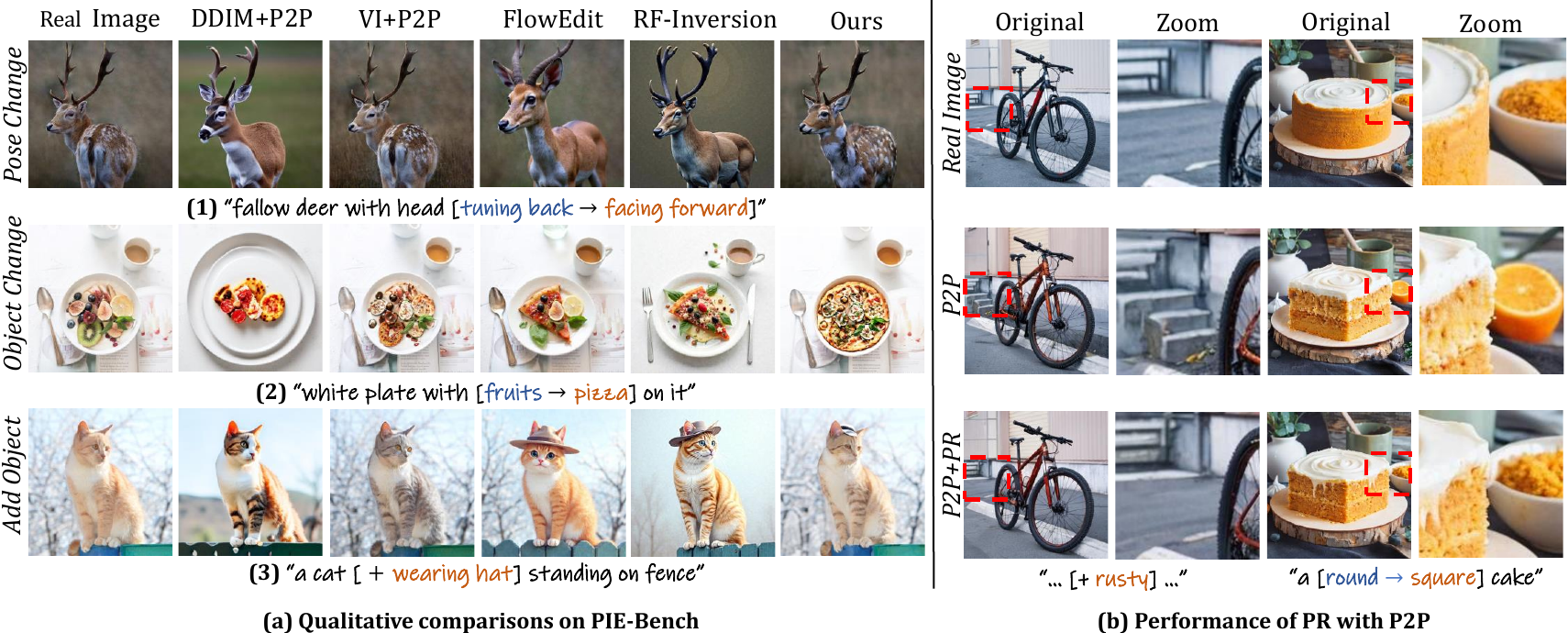}\\
      \caption{(a) Our approach achieves a superior balance between target prompt alignment and unedited region preservation. (b) The zoom column shows an enlarged view of the red-dashed-box region from the original column. PR and P2P work synergistically, with PR enhancing P2P's ability to preserve unedited regions.}
  \label{fig: quali}
\end{figure*}
\subsection{Experimental Setup}
{\bfseries Dataset. }We evaluate our method on PIE-Bench~\cite{ju2023direct}, which comprises 700 images across 4 categories (animals, people, indoor scenes, outdoor scenes) and convers 10 types of editing tasks like object addition, modification and style transfer. Each instance provides a source prompt, a target prompt and a mask indicating the edited regions. The detailed experimental setup is described in Appendix~\ref{app: metrics}.

{\bfseries Evaluation Metrics. }We assess editing fidelity through multiple measures: Structure Distance for global consistency~\cite{tumanyan2023plug}, MSE, PSNR, SSIM~\cite{wang2004image}, LPIPS~\cite{zhang2018unreasonable} for evaluating the preservation of unedited regions, and CLIPScore~\cite{hessel2021clipscore} for measuring alignment between target prompts and edited results.

{\bfseries Baselines. }  We compare \sys with representative tuning-free methods: DDIM~\cite{song2020denoising}, VI~\cite{xu2024inversion},  FlowEdit~\cite{kulikov2024flowedit} and RF-inversion~\cite{rout2024semantic}. DDIM is implemented using SD1.5~\cite{rombach2022high}, while VI adopts the latent consistency model (LCM)~\cite{luo2023latent} variant of SD1.5. FlowEdit and RF-inversion use Flux~\cite{flux2024}. \sys leverages SD1.5, Flux and LCMs from SD1.5 and SDXL1.0~\cite{podell2023sdxl}. Additionally, we evaluate P2P-enhanced~\cite{hertz2022prompt} versions of DDIM and VI, where P2P is a Unet-based~\cite{rombach2022high} attention control method. We also include StableFlow, a DiT-based~\cite{Peebles2022DiT} attention control method with Flux. Implementation details are provided in Appendix~\ref{app: implementation}.

\subsection{Quantitative and Qualitative Analysis}
We evaluate the effectiveness and efficiency of \sys in preserving consistency and producing high-quality edits.
\textbf{Quantitative Results.} 
\tabref{tab: quanti} shows that TweezeEdit performs robustly across various paradigms, including noise- (SD1.5), velocity- (Flux), and clean-image predictors (LCM), achieving superior consistency preservation while maintaining editing performance. Metric comparisons are based on mean values, with p-values computed using the Wilcoxon signed-rank test. On SD1.5, TweezeEdit outperforms DDIM in LPIPS by $-137.51$ ($p<0.01$). On Flux, TweezeEdit exceeds RF-inversion and FlowEdit across all consistency metrics while achieving higher Whole CLIPScore. Stable-flow sacrifices editability for consistency, significantly underperforming in Whole and Edited CLIPScore than TweezeEdit ($-1.25$, $p<0.01$ and $-1.34$, $p<0.01$, respectively). On LCM, even when VI is augmented with P2P for consistency enhancement, \sys (without P2P) surpasses it in both consistency and alignment (PSNR: $+2.8$, $p<0.01$; Edited CLIPScore: $+0.59$, $p<0.01$). Our method performs notably better on LCM with reduced inference steps (only 12 steps) and self-consistency, surpassing its performance on Flux. Compared to SD1.5, it achieves significant consistency improvement (MSE: $-29.19$, $p<0.01$) with minimal editing cost (Edited CLIPScore: $-0.15$, n.s.). Integrating P2P further improves our method’s consistency at a minor trade-off in alignment. Beyond these models, \sys's architecture-agnostic design enables seamless adaptation to SDXL1.0 to achieve superior alignment.

{\bfseries Qualitative evaluation. }As shown in \figref{fig: quali}, our method effectively edits images while preserving source fidelity. For examples, (1) it changes a deer's pose from backward to forward while preserving its identity characteristics (\figref{fig: quali} (a-1)); (2) it replaces fruit with pizza without altering background (\figref{fig: quali} (a-2)); and (3) it adds a hat to a cat while maintaining its original pose and visual attributes (\figref{fig: quali} (a-3)). In comparison, DDIM struggles with error accumulation and over-alignment to the target prompt. VI+P2P achieves consistency preservation but fails in editing tasks, such as missing the hat in \figref{fig: quali}. FlowEdit and RF-Inversion introduce unintended artifacts, modifying backgrounds (\figref{fig: quali} (a-2, a-3)) or altering character attributes (\figref{fig: quali} (a-1, a-3)). More visual comparison can be found in Appendix~\ref{app: w/o p2p}.

\begin{figure*}[tt]
    \centering
  \includegraphics[width=\textwidth]{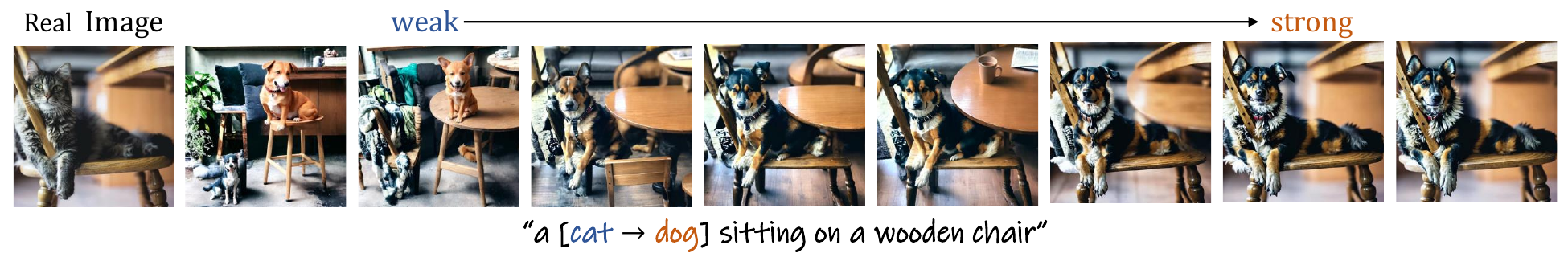}\\
    \raggedright
  \caption{\sys performance across path regularization strengths. Left to right: increasing strength enhances source consistency, where an appropriate intensity strikes a balance between consistency and target alignment.}
  \label{fig: alpha_schedule}
\end{figure*}

\subsection{Path Regularization Analysis}
Path regularization is pivotal to \sys, balancing image consistency preservation with flexible editing. We analyze its impact through three dimensions: early-step regularization, integration with attention-based methods and balance between consistency and target-alignment. Additionally, the experiments in Appendix~\ref{app: robust} confirm that path regularization remains robust to random starts and slight gradient strength perturbations.
\begin{table}
\centering
  \begin{tabular}{l c c c c c}
    \toprule
    {\multirow{2}{*}{Metrics}} & \multicolumn{5}{c}{Path Regularization Steps (\#)} \\
    
      &0 & 2 & 4 & 6 & 8\\
    \midrule
    SD$\uparrow$  & 82.1 & 57.96 & 34.25 & 17.36 & \textbf{9.71}\\
    PSNR$\uparrow$& 15.84 & 17.82& 21.03 & 24.62 & \textbf{27.19}  \\
    LPIPS$_{10^3}$$\downarrow$ & 231.04 & 183.33 & 125.30 & 81.90 & \textbf{60.23}\\    
    MSE$_{ 10^4}$$\downarrow$ & 345.00 & 222.09& 113.37 & 54.42 & \textbf{32.42} \\
    SSIM$_{10^2}$$\uparrow$ & 64.96 & 69.99 & 75.69 & 80.40 & \textbf{82.89}\\
    \midrule
    Whole$\uparrow$ & \textbf{27.08} & 26.78 & 26.43 & 25.54 & 24.22 \\
    Edited$\uparrow$& \textbf{23.76} & 23.77 & 23.27 & 22.30 & 21.15\\
    \bottomrule

\end{tabular}
  \caption{Performance across different early steps with path regularization (total steps: 12; bold values denote best results). Increasing steps boost consistency but reduce goal alignment, with an optimal tradeoff at 6 steps.
}
  \label{tab: alpha_step}
\end{table}

\textbf{Early-step regularization.} In diffusion models, early generation steps play a crucial role in shaping image structure. We thus restrict path regularization to the first $m$ steps of the 12-step process. As shown in \tabref{tab: alpha_step}, increasing $m$ enhances consistency in unedited regions but gradually reduces editing capability. To strike an optimal balance, applying path regularization to half the steps (6 of 12) preserves source image fidelity while avoiding excessive editing constraints on later steps.

{\textbf{Synergy with attention-based methods.}  Path regularization is compatible with attention-control frameworks. \figref{fig: quali} (b) demonstrates that the combination of path regularization and P2P substantially improves output consistency while reducing artifacts in standalone P2P. 

{\textbf{Balancing consistency and alignment.} Path regularization empowers users to calibrate the trade-off between consistency and text alignment. As illustrated in \figref{fig: alpha_schedule}, increasing regularization strength shifts outputs from strict text alignment to structural preservation (e.g., retaining background when converting a cat to a dog). This tunability supports diverse customization needs without architectural modifications, making \sys adaptable to both consistency-focused and creativity-driven editing scenarios.

\subsection{Perceptual Quality Assessments}
We evaluate the visual quality of the edited images using IR~\cite{xu2023imagereward}, HPSv2~\cite{wu2023human}, PickScore~\cite{kirstain2023pick}, and AES~\cite{schuhmann2022laion}. Our comparison includes TweezeEdit, FlowEdit, RF-Inversion, and Stable Flow based on Flux. Our method, TweezeEdit, achieves the best performance across all metrics. For example, its IR score is 3.81 higher than that of the second-best method. Detailed results can be found in Appendix~\ref{app: perce}.

\subsection{Text-based Translation Editing}

\begin{wrapfigure}{r}{0.48\textwidth}
    \centering
    
  \includegraphics[width=0.48\textwidth]{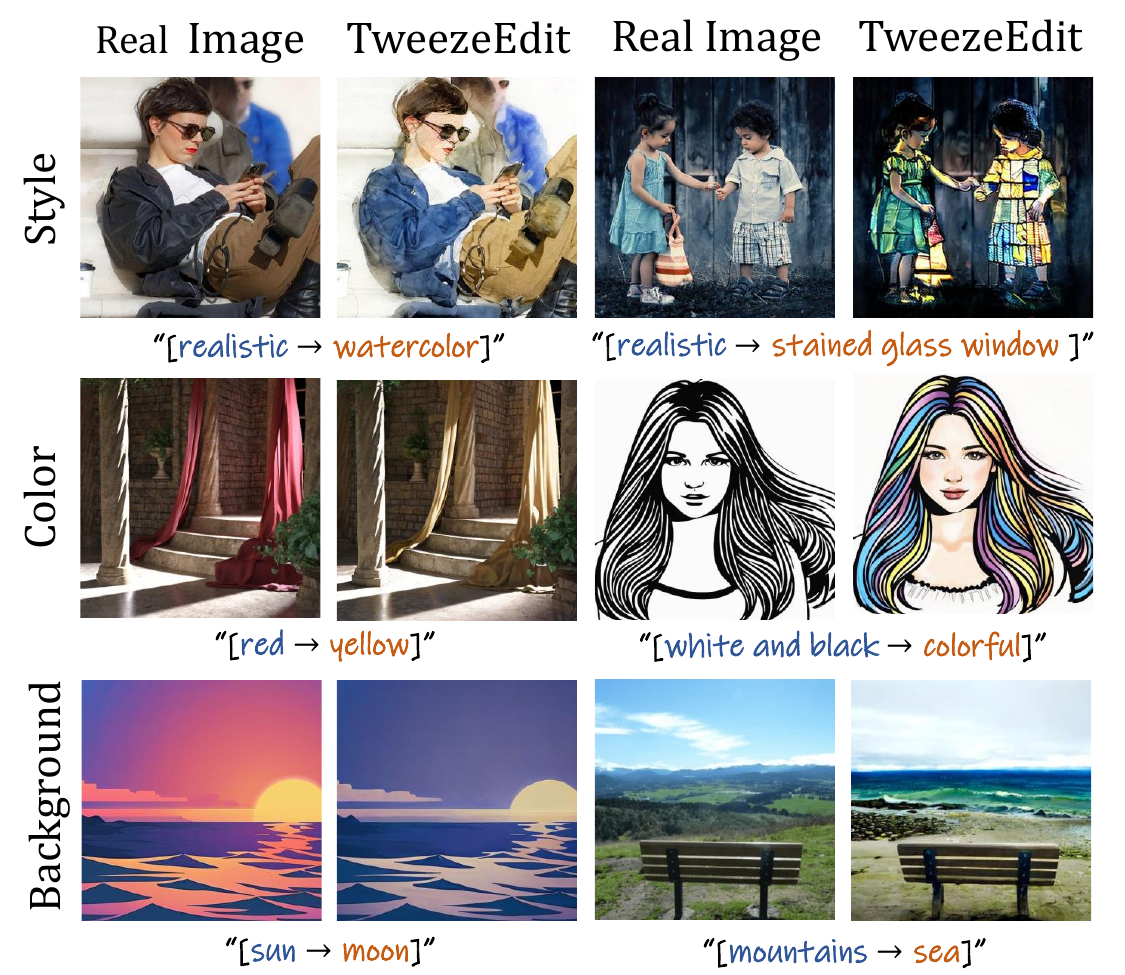}
      \caption{\sys 's performance in consistency-critical translation tasks.}
    \label{fig: app}
    \vspace{-10pt}
\end{wrapfigure}
Text-based translation editing tasks require strong consistency preservation, especially when making substantial visual modifications. As demonstrated in \figref{fig: app}, \sys effectively maintains semantic consistency in translation tasks.

\section{Conclusion}
In this work, we addresses the critical challenge of semantic fidelity loss in text-driven image editing with diffusion models, where existing methods over-align to target prompts and fail to preserve source content due to inversion inaccuracies and unconstrained denoising. Our framework, \textit{TweezeEdit}, introduces an inversion-free paradigm that regularizes the denoising trajectory between source and target images, enabling efficient editing via consistency models. By constraining divergence to prompt-relevant regions and using gradient-guided updates, TweezeEdit achieves source-consistent, target-aligned edits without architectural changes. Extensive evaluations demonstrate its superior effectiveness and efficiency across diverse tasks, highlighting the potential of path regularization to bridge the gap between creative intent and generative model limitations.

\bibliography{references}  
\bibliographystyle{references}

\newpage
\section*{Appendix}
\appendix

\section{Method}
\subsection{Assumption of Shared Noise}
\label{app: shared}

In \secref{sec: conmodel} of the main text, we assume that $z_t^{src}$ and $z_t^{tar}$ share identical noise. Theoretically, this assumption minimizes the upper bound of the distance between $z_t^{src}$ and $z_t^{tar}$ by eliminating the random terms in \eqref{eq: tmix}. Empirically, it reduces the differences caused by randomness, which would otherwise produce noisy results as shown in \figref{fig: share}.

\subsection{Path Regularization Gradient}
\label{app: gradient}
\begin{wrapfigure}{r}{0.48\textwidth}
\vspace{-10pt}
    \centering
  \includegraphics[width=0.48\textwidth]{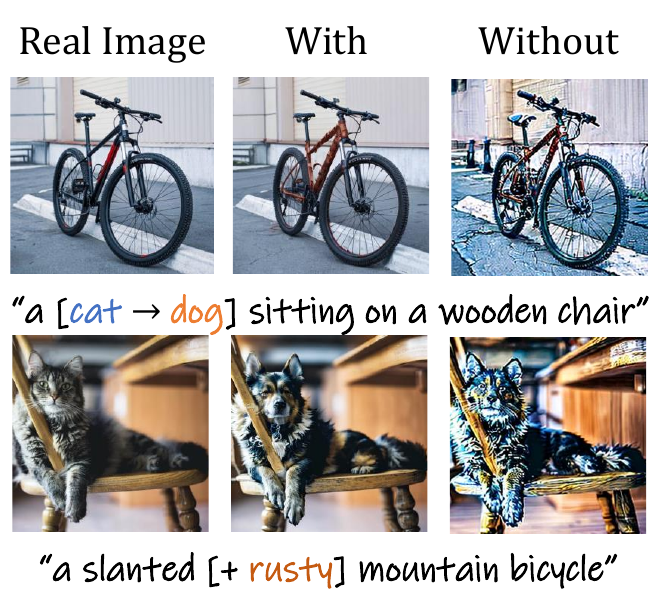}\\
    \caption{Comparison of edited results with and without shared noise}
  \label{fig: share}
  \vspace{-30pt}
\end{wrapfigure}
To approximate the integral over continuous time, we apply the integral mean value theorem with Taylor expansion at $t$: For some $\Delta_t \in [0,1]$,
\begin{equation}
\begin{aligned}
    R_t & = \gamma_t \lVert z_{t-\Delta_t}^{src} - z_{t-\Delta_t}^{tar} \rVert_2^2 \\
    & \approx \gamma_t \lVert z_t^{src} - z_t^{tar} - \Delta_t (\dot{z}_t^{src} - \dot{z}_t^{tar}) \rVert_2^2.
\end{aligned}  
\end{equation}

While the theorem does not strictly hold in high dimensions, it remains approximately valid for nearly straight diffusion model trajectories~\cite{zhou2024fast}, while the self-consistency of the consistency model further reduces the impact of this approximation.

From the relationship $z_t^{src} - z_t^{tar} = \sqrt{\bar{\alpha}_t}(z_0^{src} - z_0^{tar})$, we derive $\dot{z}_{t}^{src} - \dot{z}_{t}^{tar} = \frac{\dot{\bar{\alpha}}_t}{2\sqrt{\bar{\alpha}_t}}(z_0^{src} - z_0^{tar}),$ where $z_0^{src/tar}$ can be approximated by $f(z_{t}^{src/tar}, t)$. Substituting this yields  
\begin{equation}
    R_t \approx \gamma_t \lVert z_t^{src} - z_t^{tar} - \frac{\Delta_t \dot{\bar{\alpha}}_t}{2\sqrt{\bar{\alpha}_t}} \big(f(z_t^{src}, t) - f(z_t^{tar}, t)\big) \rVert_2^2.
\end{equation}

As $z_t^{mix}$ is updated along the direct path, we regularize its update using the gradient of $R_t$ w.r.t. $z_t^{mix}$. Note that $z_t^{src}$ (noised from $z_0^{src}$) is independent of $z_t^{mix}$, giving $\frac{\partial z_t^{src}}{\partial z_t^{mix}} = 0$. From $z_t^{tar} = z_t^{mix} - z_0^{src} + z_t^{src}$ and the approximation $z_t^{tar} \approx \sqrt{\bar{\alpha}_t}f(z_t^{tar},t) + \sqrt{1-\bar{\alpha}_t}\varepsilon$, we obtain $\frac{\partial z_t^{tar}}{\partial z_t^{mix}} = I$ and $\frac{\partial f(z_t^{tar},t)}{\partial z_t^{mix}} = \frac{I}{\sqrt{\bar{\alpha}_t}}$. Consequently, the gradient of $R_t$ is expressed as:
\begin{equation}
\label{eq: grad3}
\fontsize{8}{9} \nabla_{z_t^{mix}} R_t \approx  \hat{\gamma}_t \left[z_t^{src} - z_t^{tar} - \frac{\dot{\bar{\alpha}}_t}{4\sqrt{\bar{\alpha}_t}} (f(z_t^{src}, t) - f(z_t^{tar}, t))\right],
\end{equation}
\subsection{Selection of $\gamma_t$}
\label{app: choice}
In practice, we simplify \eqref{eq: grad3} in the main text in advance. From $z_t^{src} - z_t^{tar} = \sqrt{\bar{\alpha}_t}(z_0^{src} - z_0^{tar})$, and using \eqref{eq: approx} to approximate $z_0^{tar}$ once again. We have:
\begin{equation}
\label{eq: grad2}
\nabla_{z_t^{mix}} R_t \approx  \hat{\gamma}_t \left[(f(z_t^{src}, t) - f(z_t^{tar}, t))\right],
\end{equation}
     where we set $\Delta_t \approx \frac{1}{2}$ and define $\hat{\gamma}_t := -2\gamma_t \big(-1 + \frac{\dot{\bar{\alpha}}_t}{4\sqrt{\bar{\alpha}_t}}\big)\big(-1 + \frac{\dot{\bar{\alpha}}_t}{4{\bar{\alpha}_t}}\big)$ for simplicity.
Note that this definition of $\hat\gamma_t$ differs from the one in the main text, but we reuse the notation for clarity in this derivation. To ensure the injection of semantics, we constrain $|z_{t}^{tar}-z_{t}^{src}|\ge|z_{t-1}^{tar}-z_{t-1}^{src}|$, and thus we select $\hat{\gamma}_t$ within the range $[-(\sqrt{\bar\alpha_{t}}-\sqrt{\bar{\alpha}_{t-1}}), 0]$. When $\hat{\gamma}_t=1$, we have
\begin{equation*}
    |z_{t}^{tar}-z_{t}^{src}|=|z_{t-1}^{tar}-z_{t-1}^{src}|,
\end{equation*}
 achieving maximum image consistency preservation in this case. Alternatively, we can directly use $z_{t}^{tar}-z_{t}^{src}=z_{t-1}^{tar}-z_{t-1}^{src}$ to bypass the diffusion model's prediction, thereby saving computational overhead. On the contrary, $\hat{\gamma}_t=0$ achieves the maximum editing effect. The updated algorithm is summarized in \algref{alg2}.
\begin{algorithm}
\caption{TweezeEdit}
\label{alg2}
    \SetAlgoLined
    \SetKwInOut{Input}{Input}
    \SetKwInOut{Output}{Output}
    \Input{Source image $z_0^{src}$, source prompt $P^{src}$, target prompt $P^{tar}$, regularization scheduler $\hat{\gamma}_t$ and consistency model $f$}
    \Output{Edited image $z_0^{tar}$}
    $z_T^{mix} = z_0^{src}$ {// initialization} \; 
    \For{$t \leftarrow T$ \KwTo $1$}{
        {//Obtain samples in denoising path}\\
        Sample $\epsilon\sim\mathcal{N}(0,I)$\;
        $z_t^{src}=\sqrt{\bar{\alpha}_t}z_0^{src}+\sqrt{1-\bar{\alpha}_t}\epsilon$\;
        $z_t^{tar}=z_t^{mix}-z_0^{src}+z_t^{src}$\;
        {//Consistency model's prediction}\\
        $\hat{z}_0^{src}=f(z_t^{src},t,P^{src})$\;
        $\hat{z}_0^{tar}=f(z_t^{tar},t,P^{tar})$\;
        {//Editing direction in direct path}\\
        $v_t=z_0^{src}+\sqrt{\bar{\alpha}_{t-1}}(\hat{z}_0^{tar}-\hat{z}_0^{src})$  \; 
        {//Gradient of denoising path regularization}\\
        $\nabla_{z_t^{mix}}R_t = \hat{\gamma}_t \left[(\hat{z}_0^{src}-\hat{z}_0^{tar})\right]$  \;
        {//Update step}\\
        $z_t^{mix}=v_t-\nabla_{z_t^{mix}}R_t$  \;}
    \Return Edited image $z_0^{tar}=z_0^{mix}$
\end{algorithm}

\subsection{Application of TweezeEdit to Noise and Velocity Predictors}
\label{app: para}
The relationship between noise/velocity (linear path) predictors and clean image predictors can be expressed as:
\begin{align*}
    f_\theta(z_t,t)&\approx \frac{z_t-\sqrt{1-\bar{\alpha}_t}\epsilon_\theta(z_t,t)}{\sqrt{\bar{\alpha}_t}} \\
    f_\theta(z_t,t)&\approx z_t - t\cdot v_\theta(z_t,t)
\end{align*}
Based on this relationship, we can predict $\hat{z_0^{src}}$ and $\hat{z_0^{tar}}$ using the above formulations to update \eqref{eq: update}.

For noise predictors,
\begin{align*}
    \hat{z_0^{src}} = \frac{z_t-\sqrt{1-\bar{\alpha}_t}\epsilon_\theta(z_t,t,P^{src})}{\sqrt{\bar{\alpha}_t}}\\
    \hat{z_0^{tar}} = \frac{z_t-\sqrt{1-\bar{\alpha}_t}\epsilon_\theta(z_t,t,P^{tar})}{\sqrt{\bar{\alpha}_t}}
\end{align*}

For velocity predictors,
\begin{align*}
    \hat{z}_0^{src} = z_t - t\cdot v_\theta(z_t,t, P^{src})\\
    \hat{z}_0^{tar} = z_t - t\cdot v_\theta(z_t,t, P^{tar})
\end{align*}    
\section{Experimental Details}
\subsection{Metrics}
\label{app: metrics}
In this section, we specify the used metrics.
Five metrics are employed to quantify consistency.
\begin{itemize}
    \item \textbf{Structure Distance} ~\cite{tumanyan2023plug} is a similarity metric quantifying the structural resemblance between two images by self-similarity distances derived from the DINO-ViT model~\cite{kwon2022diffusion},  which extracts robust, semantically meaningful features through self-supervised learning.
    \item \textbf{Peak Signal-to-Noise Ratio (PSNR)} quantifies the ratio of the maximum possible signal power to the power of distorting noise to measure the quality of reconstructed images compared to the original. 

    \item \textbf{Learned Perceptual Image Patch Similarity (LPIPS)}~\cite{zhang2018unreasonable} is a perceptual quality metric that measures the similarity between two images by comparing feature activations extracted from deep neural networks.
    \item \textbf{Mean Squared Error (MSE)} evaluates the similarity of two images by measuring the average squared difference between them.
    \item \textbf{Structural Similarity Index Measure (SSIM)}~\cite{wang2004image} assesses the similarity between two images by comparing luminance, contrast, and structural information.
\end{itemize}
One metric is used to quantify alignment.
\begin{itemize}
    \item  \textbf{CLIPScore}~\cite{hessel2021clipscore} evaluates image-caption alignment by computing the cosine similarity of CLIP embedding between images and text. .whole and edited. Whole CLIPScore assesses the entire image against the text, while Edited CLIPScore focuses on the edited region of the image compared to the text. 
\end{itemize}
Four metrics are utilized to quantify image perceptual quality.
\begin{itemize}
    \item \textbf{Aesthetic Score (AES)}~\cite{schuhmann2022laion} evaluates image aesthetic quality based on an aesthetic predictor, which is trained on CLIP image embeddings in  LAION-5B dataset to predict aesthetic scores from 1 to 10, reflecting human subjective preferences.
    \item \textbf{ImageReward (IR)}~\cite{xu2023imagereward} quantifies human preference based on a general-purpose text-to-image human preference reward model (RM), which is trained on 137k pairs of expert comparisons. 
    \item \textbf{Human Preference Score v2 (HPSv2)}~\cite{wu2023human} predicts human preferences for generated images using a scoring model, which is a fine-tuned CLIP model trained on HPD v2—a dataset comprising 798,090 human preference choices across 433,760 image pairs.
    \item \textbf{PickScore (PS)}~\cite{kirstain2023pick} predicts human preferences for generated images using a CLIP-based scoring function trained on Pick-a-Pic—a large, open dataset of text-to-image prompts paired with real user preferences collected through a web app.
\end{itemize}
\subsection{Implementation Details}
\label{app: implementation}
\begin{table}[t]

\centering

\begin{tabular}{lcccc}
\toprule
Method & IR & AES & HPSv2 & PS \\
\midrule
FLowEdit (Flux) & 73.11 & 28.02 & 22.16 & 6.75  \\
RF-inversion (Flux) & 69.88 & 27.74 & 21.99 & 6.43\\
Stable Flow (Flux)& 30.43 & 27.26 & 21.37 & 6.51  \\
TweezeEdit (Flux) & 76.92 & 28.05 & 22.22 & 6.76  \\
\bottomrule
\end{tabular}
\caption{Perceptual Quality Assessment Results}
\label{tab: perce}
\end{table}
For all other methods, we maintained their official hyperparameters settings. The experimental configurations were implemented as follows: DDIM (SD1.5) uses 50 generation steps with a Classifier-Free Guidance (CFG) scale of 7.5 for both source and target prompts. DDIM(SD1.5)+P2P follows the same base parameters as DDIM while incorporating P2P-specific settings of 0.6 for self replace steps and 0.4 for cross-replace steps. Flowedit (Flux) was configured with 28 steps and an n-max value of 24, using CDF scales of 1.5 for source prompts and 5.5 for target prompts. Virtual Inversion (VI) (LCM: SD1.5) employed 12 steps (extended to 15 when combined with P2P), with CFG scales set to 1.0 for source prompts and 2.3 for target prompts. VI(LCM: SD1.5)+P2P configuration inherited base parameters from VI while incorporating setting of self replace steps 1.0 and 0.7 cross replace steps 0.7. 

For TweezeEdit, we implemented the following configurations. TweezeEdit (SD1.5) used 25 steps with CFG scales of 3.5 (source) and 7.5 (target), regularizing the first step with  $\gamma=-(\sqrt{\bar\alpha_{t}}-\sqrt{\bar{\alpha}_{t-1}})$. TweezedEdit (LCM: SD1.5) used 12 steps with equal CFG scales of 1.5 for both prompts, applying path regularization with $\gamma=-(\sqrt{\bar\alpha_{t}}-\sqrt{\bar{\alpha}_{t-1}})$ during the first 6 steps.TweezedEdit (LCM: SD1.5+P2P) maintained the same CFG, total steps with TweezeEdit (LCM: SD1.5), and applies path regularization for the first 3  steps $\gamma=-(\sqrt{\bar\alpha_{t}}-\sqrt{\bar{\alpha}_{t-1}})$, along with P2P configuration of self replace steps 0.6 and cross replace steps 0.4. TweezeEdit (Flux) runs for 28 steps with source prompt CFG 1.5 and target prompt CFG 5.5, regularizing the first 4 steps with $\gamma=-0.8(\sqrt{\bar\alpha_{t}}-\sqrt{\bar{\alpha}_{t-1}})$. Finally, TweezeEdit (LCM: SDXL1.0) runs for 15 steps with equal CFG values of both prompts, regularizing the first 3 steps with $\gamma=-(\sqrt{\bar\alpha_{t}}-\sqrt{\bar{\alpha}_{t-1}})$.

All experiments were conducted on a single NVIDIA RTX A6000 GPU (48GB) with Ubuntu 22.04.5 LTS as the operating system. Each editing task was executed with a single run of the algorithm, and the results were recorded for evaluation.

\section{Quantitative Results}

\subsection{Results of perceptual quality assessment}
\label{app: perce}

We evaluate the perceptual quality of TweezeEdit, FlowEdit, RF-Inversion, and Stable Flow based on Flux. As shown in \tabref{tab: perce}, our method achieves the best visual quality across multiple metrics, including IR~\cite{xu2023imagereward}, AES~\cite{schuhmann2022laion}, HPSv2~\cite{wu2023human}, and PS~\cite{kirstain2023pick}. This demonstrates that our editing method ensures high-quality outputs by effectively balancing consistency and alignment.

\subsection{Runtime Efficiency of TweezeEdit}
\begin{table}[t]
    \centering
    \begin{tabular}{c c c c}
        \toprule
        Method & Steps & path regularization &  time (s)\\
        \midrule
         TweezeEdit(SD1.5) & 12 & 6  & 1.63$\pm{0.01}$\\
         TweezeEdit(SD1.5)+P2P& 12 & 3 & 2.49$\pm{0.01}$\\
         TweezeEdit(SDXL) & 15 & 3 & 2.27$\pm{0.01}$\\
     \bottomrule
    \end{tabular}
    \caption{Efficiency of TweezeEdit}
    \label{tab: efficiency}
\end{table}
As shown in \tabref{tab: efficiency}, TweezeEdit efficiently processes images, demonstrating its suitability for real-time applications. Here, we use bypass trick in Appendix~\ref{app: gradient} to saving computational overhead.

\subsection{Robustness Analysis of Path Regularization}
\label{app: robust}
\begin{table*}[tt]

\centering

\begin{tabular}{lccccccc}
\hline
Exp & Distance ($10^3$) & PSNR & LPIPS ($10^3$) & MSE ($10^4$) & SSIM ($10^2$) & Whole & Edited \\ 
\hline
SV & $17.62 \pm 0.24$ & $24.59 \pm 0.08$ & $82.50 \pm 0.59$ & $54.42 \pm 0.69$ & $80.34 \pm 0.10$ & $25.43 \pm 0.10$ & $22.30 \pm 0.05$ \\
\hline
GP & $17.81 \pm 0.05$ & $24.77 \pm 0.40$ & $83.33 \pm 0.14$ & $55.36 \pm 0.15$ & $80.24 \pm 0.01$ & $25.48 \pm 0.01$ & $22.36 \pm 0.01$ \\
\hline
\end{tabular}
\caption{Sensitivity experiment results}
\label{tab: robust}
\end{table*}
We empirically validate the robustness of our path regularization through two key experiments (\tabref{tab: robust}): (1) Seed Variation (SV): When varying the random initialization, the results exhibit negligible fluctuations, demonstrating that our method (Eq.11) maintains consistent anchor points across different starting conditions. (2) Gradient Perturbation (GP): Introducing Gaussian noise to the gradients leads to minimal performance variation, confirming insensitivity to small optimization disturbances. Together, these results underscore that our approach remains stable under both stochastic initialization and gradient noise.

\section{Qualitative Results}
\subsection{Visual Results of TweezeEdit without P2P}
\label{app: w/o p2p}
In this section, we present a visual comparison of editing results to evaluate TweezeEdit (LCM: SD1.5 without P2P) against FlowEdit (Flux), RF-Inversion (Flux), and StableFlow (Flux). For this experiment, TweezeEdit applies path regularization during the first 10 steps of its 15-step generation process. The qualitative results demonstrate that TweezeEdit better preserves the original content while maintaining text-aligned editing, compared to the baseline methods—even though they use the advanced base model Flux.

\newpage
\begin{figure*}[tt]
    \centering
    \includegraphics[width=0.95\linewidth]{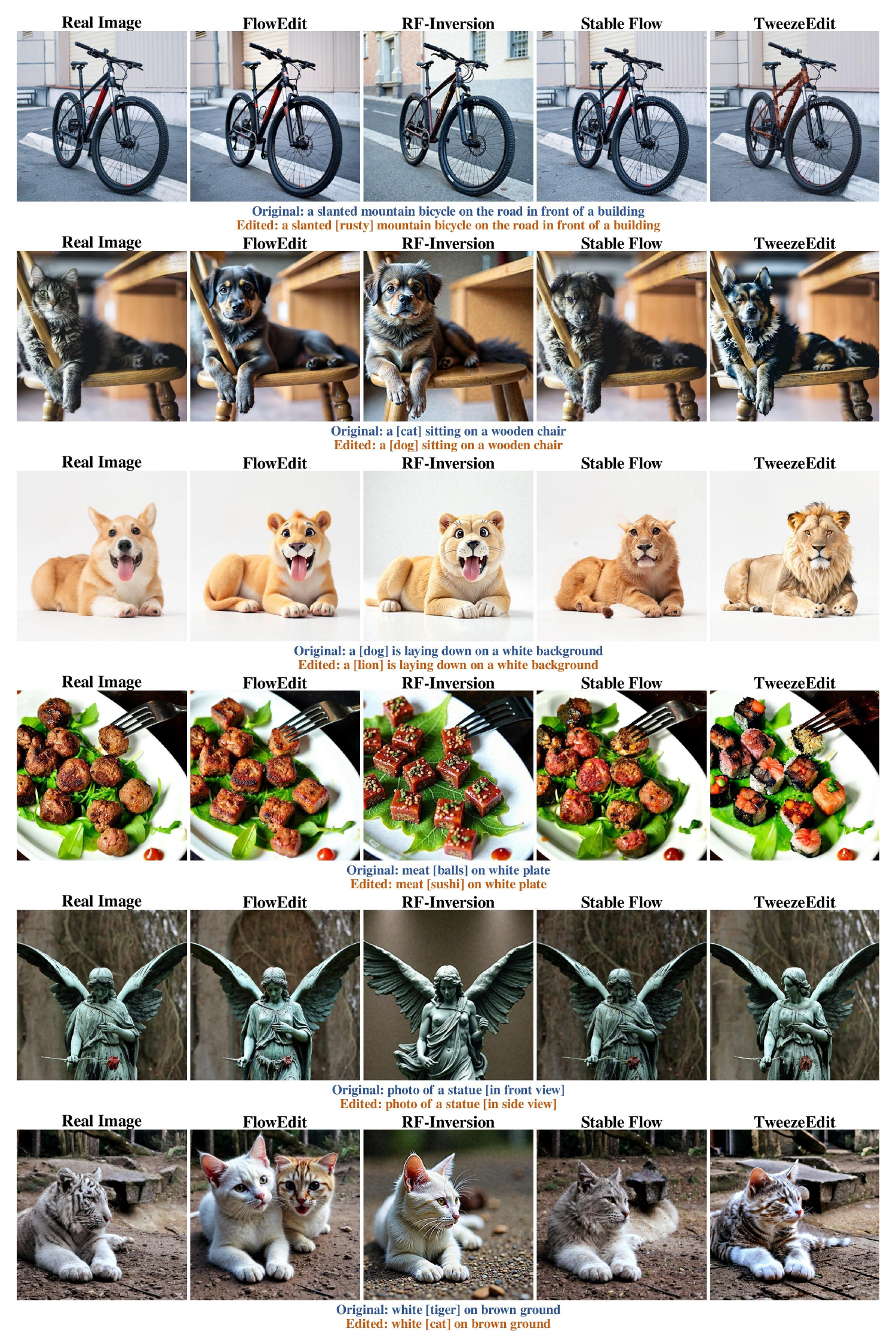}
\end{figure*}
\begin{figure*}[tt]
    \centering
    \includegraphics[width=0.95\linewidth]{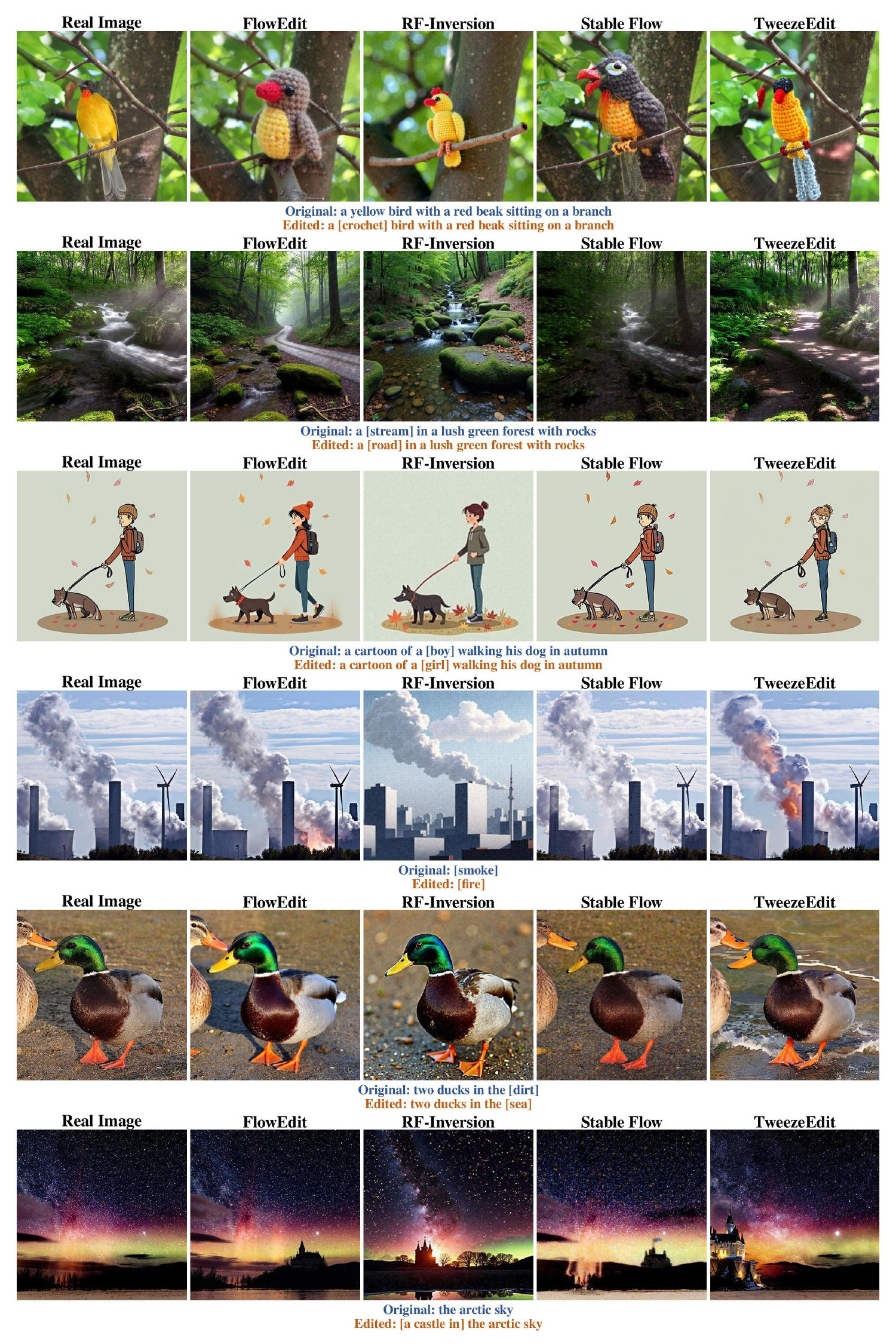}
\end{figure*}
\begin{figure*}[tt]
    \centering
    \includegraphics[width=0.95\linewidth]{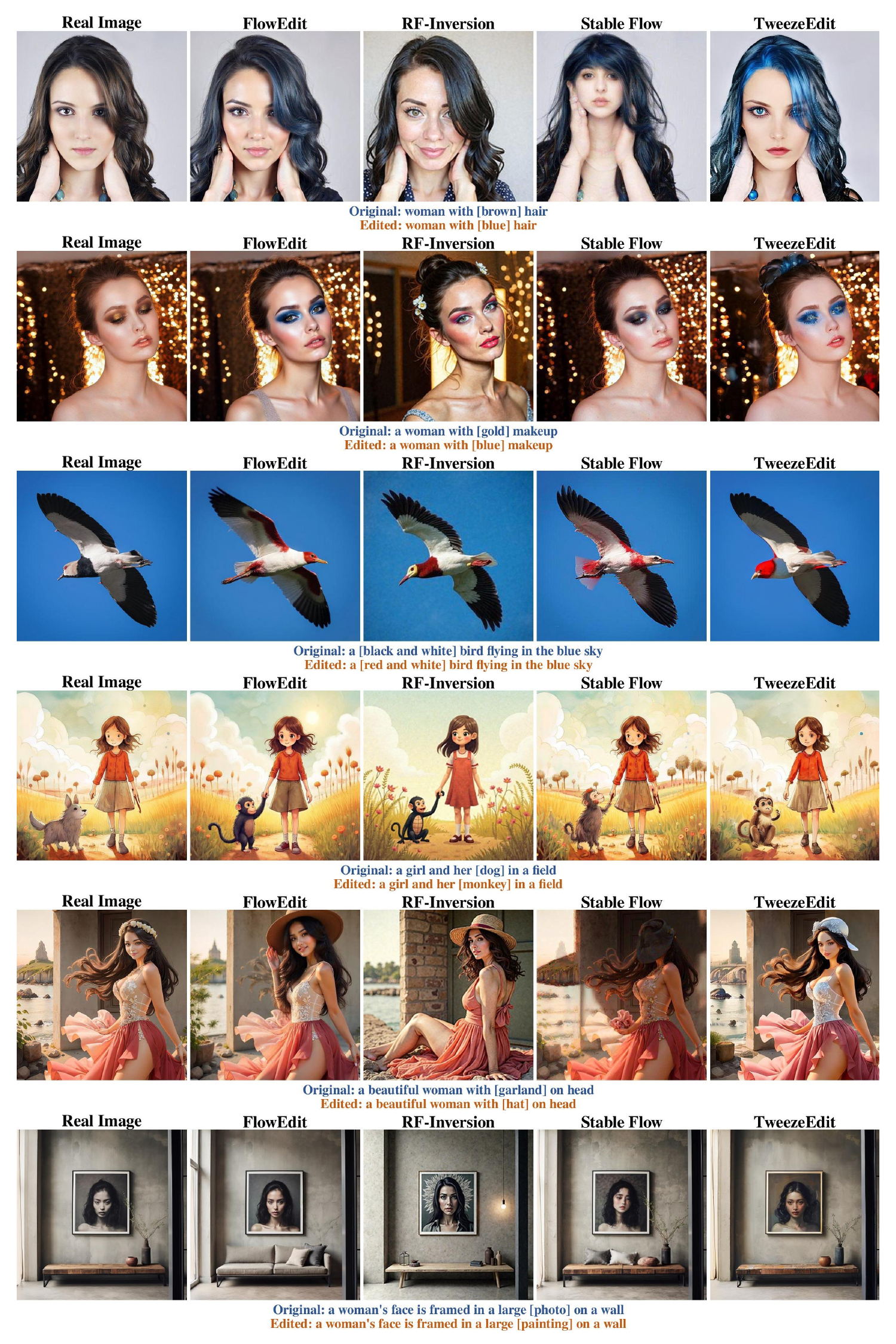}
\end{figure*}
\begin{figure*}[tt]
    \centering
    \includegraphics[width=0.95\linewidth]{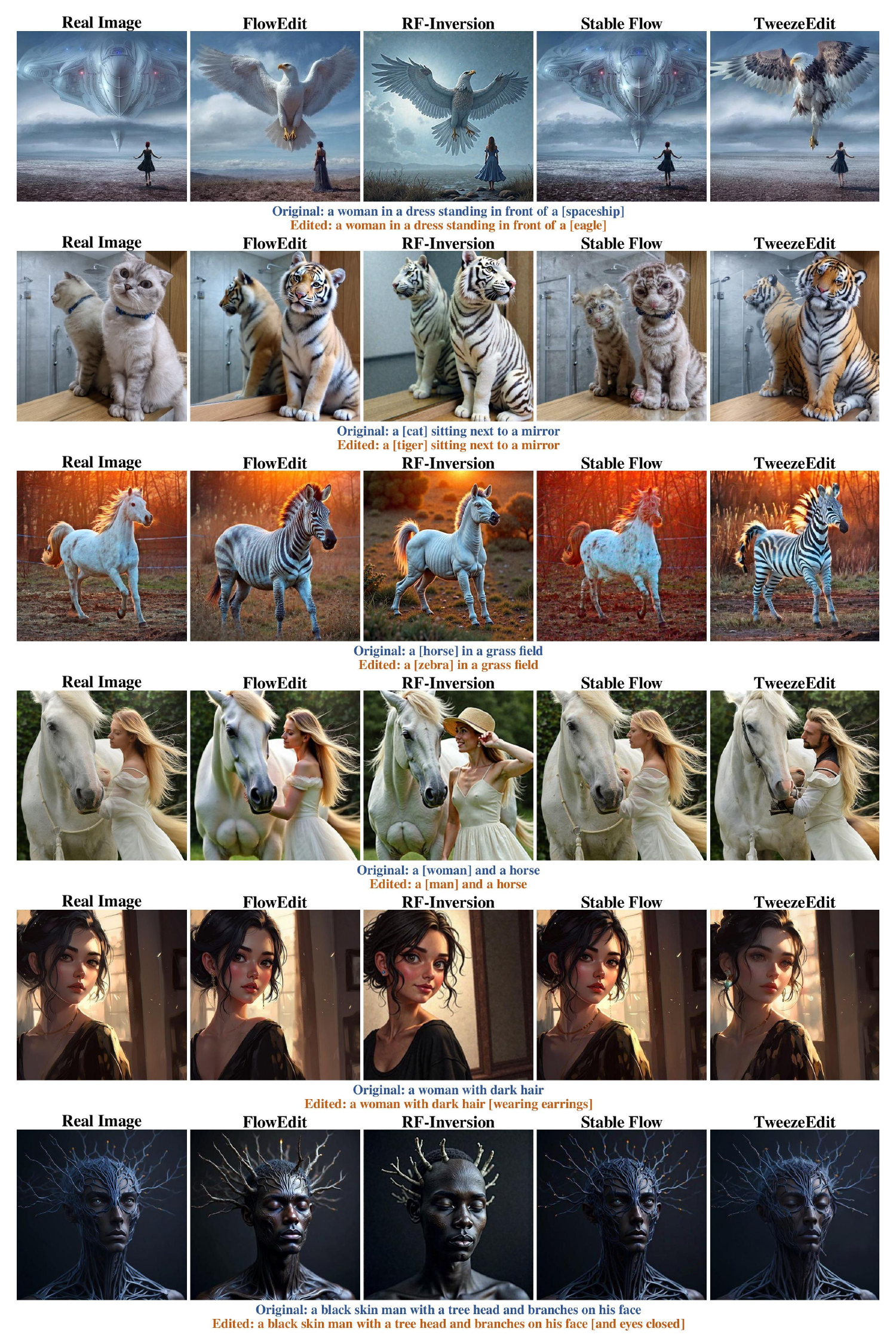}
\end{figure*}
\begin{figure*}[tt]
    \centering
    \includegraphics[width=0.95\linewidth]{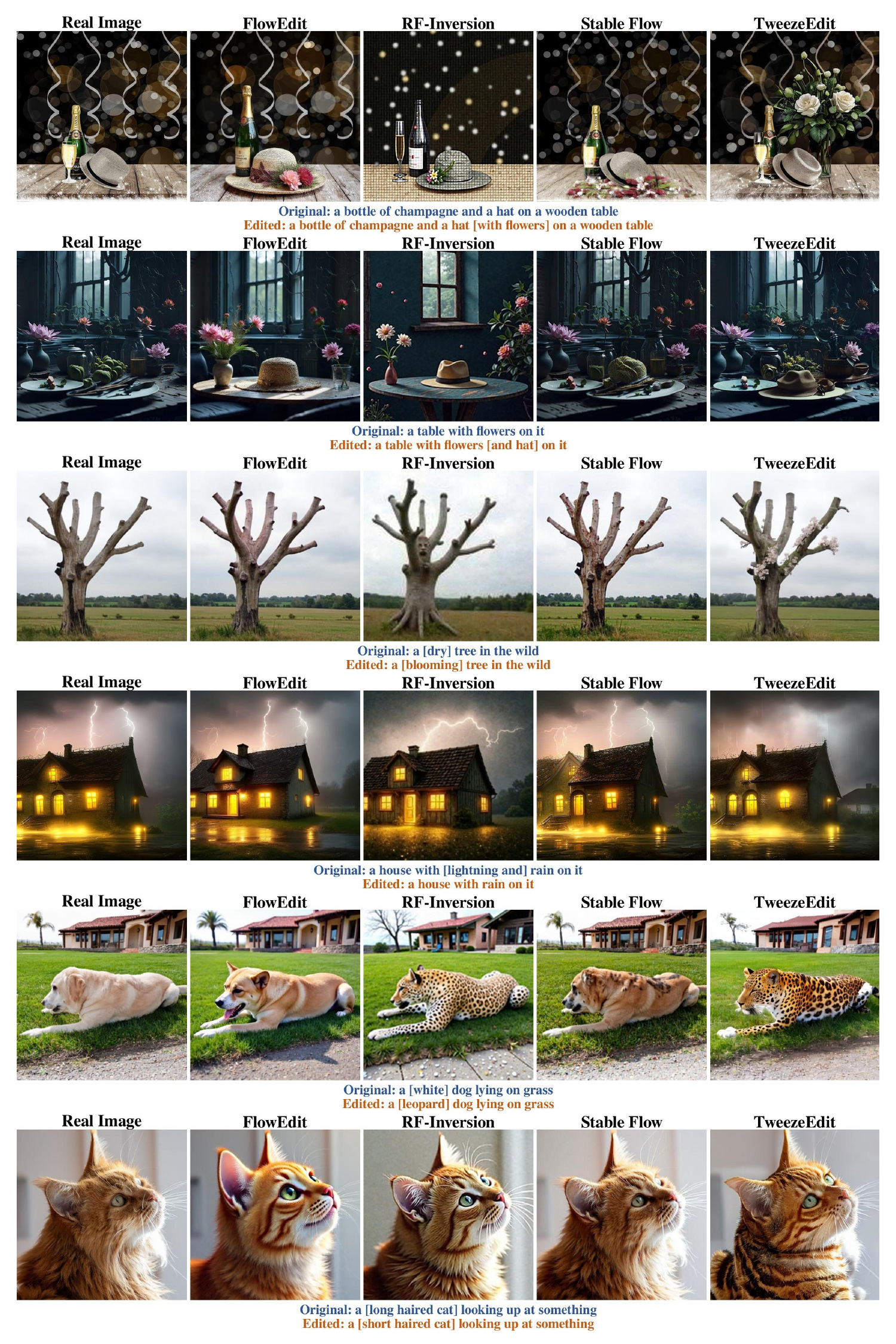}
\end{figure*}
\begin{figure*}[tt]
    \centering
    \includegraphics[width=0.95\linewidth]{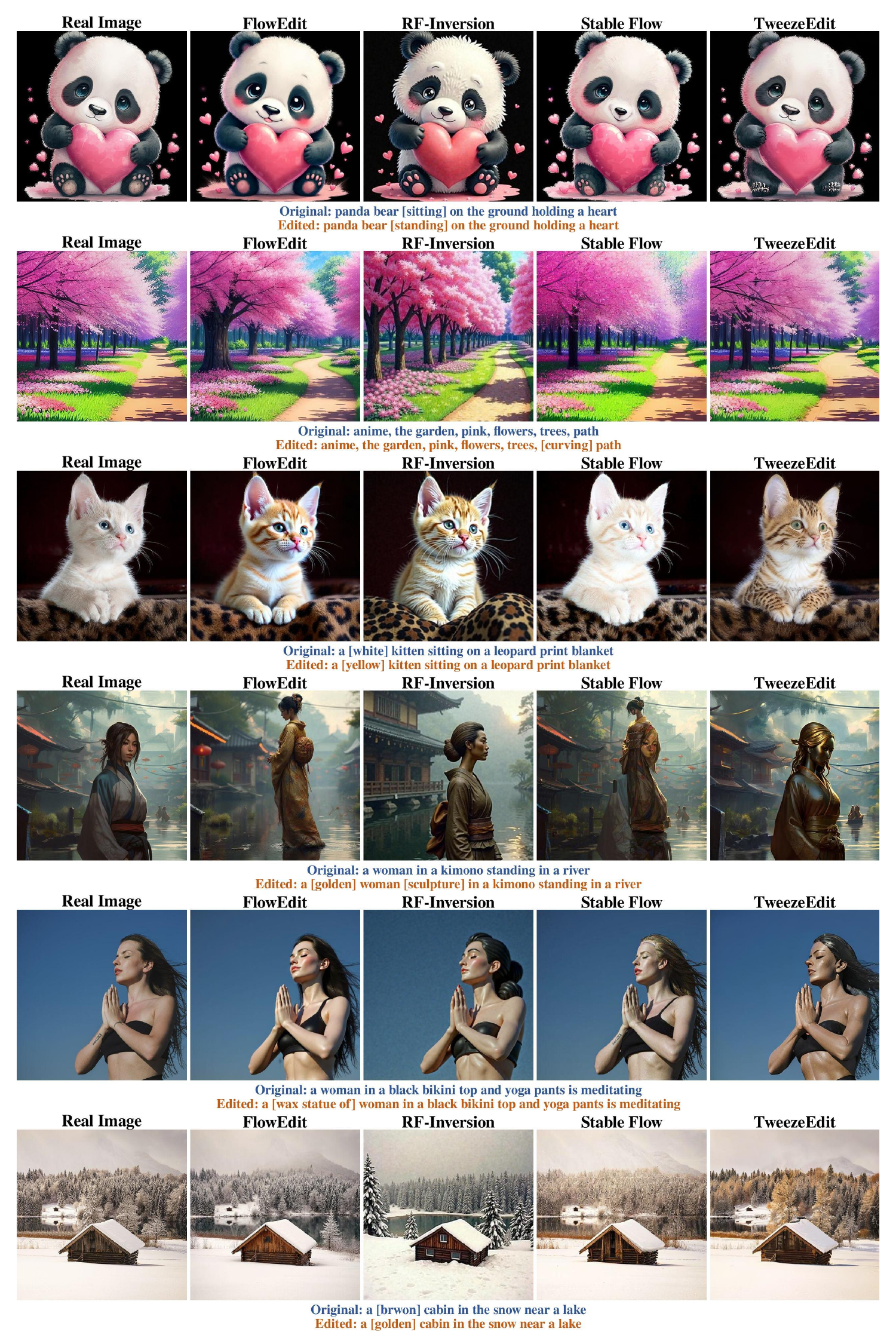}
\end{figure*}
\end{document}